\documentclass[10pt,twocolumn,letterpaper]{article}

\usepackage{cvpr}              % camera-ready
\definecolor{cvprblue}{rgb}{0.21,0.49,0.74}
\usepackage[breaklinks,colorlinks,allcolors=cvprblue]{hyperref}

\def\paperID{VGBE-13}
\def\confName{CVPR}
\def\confYear{2026}

\title{Risk-Controllable Multi-View Diffusion for Driving Scenario Generation}

\author{
Hongyi Lin$^{1,2}$ \quad Wenxiu Shi$^{3}$ \quad Heye Huang$^{4,*}$ \quad Dingyi Zhuang$^{2,*}$\\
Song Zhang$^{5}$ \quad Yang Liu$^{1}$ \quad Xiaobo Qu$^{1}$ \quad Jinhua Zhao$^{2}$\\[0.5em]
{
$^{1}$ Tsinghua University \quad
$^{2}$ MIT \quad
$^{3}$ Z-One Tech \quad
$^{4}$ KAIST \quad
$^{5}$ Chengdu Tianfu Invo Tech
}\\[0.5em]
{\tt\small \{heyeh,dingyi\}@mit.edu}\\
}

\begin{document}
\maketitle

\begin{abstract}
Generating safety-critical driving scenarios is crucial for evaluating and improving autonomous driving systems, but long-tail risky situations are rarely observed in real-world data and difficult to specify through manual scenario design. Existing generative approaches typically treat risk as an after-the-fact label and struggle to maintain geometric consistency in multi-view driving scenes. We present RiskMV-DPO, a general and systematic pipeline for physically-informed, risk-controllable multi-view scenario generation. By integrating target risk levels with physically-grounded risk modeling, we synthesize diverse and high-stakes dynamic trajectories that serve as explicit geometric anchors for a diffusion-based video generator. To ensure spatial-temporal coherence and geometric fidelity, we introduce a geometry-appearance alignment module and a region-aware direct preference optimization (RA-DPO) strategy with motion-aware masking to focus learning on localized dynamic regions. Experiments on the nuScenes dataset show that RiskMV-DPO can freely generate a wide spectrum of diverse scenarios while maintaining visual quality, improving 3D detection mAP from 18.17 to 30.50 and reducing FID to 15.70. Our work shifts the role of world models from passive environment prediction to proactive, risk-controllable synthesis, providing a scalable toolchain for the development of embodied intelligence. 
\end{abstract}
\section{Introduction}
\label{sec:intro}

Autonomous driving systems are ultimately judged by how safely and reliably they behave in the open world, including rare but safety-critical events \cite{zhang2026world,he2025dual,fei2024critical}. Since real-world logs under-sample these cases, driving scenario generation has become essential for data augmentation, stress testing, and systematic safety evaluation \cite{lin2023generative,wang2025safety}. With production vehicles relying heavily on camera suites, generating temporally coherent, multi-view consistent driving sequences is particularly valuable \cite{zhang2025can, lin2025high}.

Despite rapid progress, generating long-tail risky scenarios remains a significant bottleneck \cite{he2025generative}. A common practice is to generate a large pool of driving scenes and then mine rare events, but truly high-risk situations occur with extremely low probability \cite{zhang2024chatscene}. Another practice is to handcraft scenarios by specifying parameters such as near-collisions or road intrusions \cite{xu2025diffscene}. While offering direct control, these "sampling-and-filtering" or manual design approaches inevitably reflect human priors and lack the efficiency to synthesize the vast, non-intuitive space of hazardous interactions. What is missing is a general pipeline to transform "risk" from an after-the-fact label into a proactive, time-resolved control signal that can autonomously guide the synthesis of diverse long-tail hazards.

Meanwhile, the dynamic nature of driving scenes introduces a second bottleneck: geometry and robustness \cite{cai2026text2scenario}. While diffusion models have advanced visual realism, driving applications demand more. They require cross-view metric-level consistency, motion continuity, and stability under occlusion and lighting changes. For consistency, many attempts rely on large-scale question-answer datasets to teach VLMs spatial concepts, but this supervision rarely provides a solid geometric prior that enforces physically grounded alignment across cameras \cite{wang2026vggdrive, qu2023envisioning}. This motivates mechanisms that inject 3D priors into diffusion training and align the model’s capacity with localized, motion-dominant regions that matter most for driving risk.

In short, these challenges are twofold: \textit{making risk controllable} for driving scenario generation, and \textit{making multi-view geometry reliable} under outdoor dynamic conditions. Therefore, we propose risk-controllable multi-view diffusion for driving scenario generation (RiskMV-DPO), a framework that makes risk a first-class control knob while improving geometric fidelity and localized dynamic realism. The contributions can be summarized as follows:

\begin{itemize}
\item We propose a general pipeline for risk-controllable synthesis that reformulates driving risk as a proactive control signal, enabling the autonomous generation of dynamic trajectories and 3D boxes at target risk levels.
\item We introduce a geometry-appearance alignment module that injects compact 3D priors into multi-view diffusion to ensure metric-level cross-view consistency and geometric plausibility.
\item We develop multi-view consistent RA-DPO with motion-aware masking to align scene synthesis with localized preferences in dynamic regions, significantly enhancing temporal and motion realism.
\end{itemize}

%-------------------------------------------------------------------------

\section{Related Work}
\label{sec:related_work}

%-------------------------------------------------------------------------
\subsection{Driving scene and multi-view video generation}

Driving scenario generation has evolved from neural simulators to large-scale world models~\cite{zhang2026dynamic, lin2025big}. DriveGAN~\cite{kim2021drivegan} learns an end-to-end simulator conditioned on ego actions, while GAIA-1~\cite{hu2023gaia} models driving as token sequences with video, text, and action conditioning. Diffusion-based world models, such as DriveDreamer~\cite{wang2024drivedreamer} and DriveDreamer-2~\cite{zhao2025drivedreamer}, further improve realism and controllability. 

For multi-view driving videos, explicit structural controls are often introduced to enhance cross-view consistency. MagicDrive~\cite{gao2023magicdrive} conditions on camera poses, maps, and 3D boxes with cross-view attention. DrivingDiffusion~\cite{li2024drivingdiffusion} generates multi-view videos from 3D layouts and enforces both cross-view and cross-frame consistency. MagicDrive-V2~\cite{gao2025magicdrive} extends this toward longer and higher-resolution generation, and recent work explores stronger spatiotemporal-view attention~\cite{lu2024seeing}. However, explicitly controlling long-tail risk remains difficult, and multi-view geometry is still fragile in dynamic outdoor scenes.

%-------------------------------------------------------------------------
\subsection{Risk modeling and risk-aware simulation}

Risk modeling often relies on trajectory-based criticality metrics, such as TTC variants~\cite{xu2023deep,westhofen2023criticality}, and formal safety envelopes such as RSS~\cite{shalev2017formal}. Recent work has moved toward risk forecasting under interaction and uncertainty. For example, RiskNet~\cite{liu2026risknet} combines field-theoretic interactions with multi-modal trajectory prediction for time-evolving risk estimation, while SafeDrive~\cite{zhou2026safedrive} quantifies perceived risk via risk fields to guide an LLM driving agent.

Risk-aware testing frameworks use risk signals to search for rare failures more efficiently. Scenic~\cite{fremont2019scenic} specifies scenarios as constrained distributions, VerifAI~\cite{dreossi2019verifai} supports simulation-based falsification and parameter synthesis, and Adaptive Stress Testing~\cite{koren2018adaptive} uses RL to discover failure trajectories. Despite these advances, most methods still use risk as a threshold or post-hoc analysis signal, rather than as a controllable and time-resolved target for generation.

%-------------------------------------------------------------------------
\subsection{3D priors for multi-view generation}

Recent multi-view diffusion methods increasingly incorporate explicit 3D signals to reduce cross-view drift and improve structural consistency~\cite{tang2025priorfusion}. In driving, MagicDrive~\cite{gao2023magicdrive} and DrivingDiffusion~\cite{li2024drivingdiffusion} condition on 3D controls and cross-view structure, while DriveScape~\cite{wu2025drivescape} scales multi-view synthesis with 3D-guided feature fusion. However, existing conditioning often remains brittle under occlusion, fast motion, and complex outdoor dynamics. This motivates our geometry-appearance alignment, which injects compact 3D priors into diffusion training.

%-------------------------------------------------------------------------
\subsection{DPO for generative models}

DPO has become a practical paradigm for aligning generative models with preference data. Diffusion-DPO~\cite{wallace2024diffusion} extends DPO to diffusion models, and DSPO~\cite{zhu2025dspo} improves objective matching under score-based training. For video generation, DenseDPO~\cite{wu2025densedpo} motivates finer temporal preference signals, while LocalDPO~\cite{huang2026mind} introduces region-level mask-guided preference optimization. However, existing preference alignment methods are rarely designed for multi-view scenarios, which motivates our multi-view motion-aware masking and region-aware DPO design.
\section{Problem Formulation}

We study multi-view driving scenario generation under a target risk condition. Let $\mathcal{V}=\{1,\dots,V\}$ denote the camera set, and $\mathbf{I}_{1:T}=\{I_{1:T}^{v}\}_{v\in\mathcal{V}}$ denote the observed multi-view history. Given structured context, including HD maps $\mathbf{M}$ and text $\mathbf{y}$, the goal is to generate future video at a user-specified risk level, which can be either a scalar $r^{*}$ or a time-resolved profile $\boldsymbol{r}^{*}_{1:H}$ over a future horizon $H$.

Our key idea is to treat risk as a time-resolved conditioning signal, generate trajectories and 3D boxes at the target risk, and use them as structured controls for multi-view diffusion. Let $\mathcal{A}$ denote the set of traffic participants. For each agent $a\in\mathcal{A}$, we define the motion control signal as
\begin{equation}
\mathbf{U}=\Big\{\big(\mathbf{x}^{a}_{1:H},\mathbf{b}^{a}_{1:H}\big)\Big\}_{a\in\mathcal{A}} .
\label{eq:motion-control-signal}
\end{equation}

The generation process is decomposed into two stages. First, a risk control module $g(\cdot)$ predicts physically grounded motion conditioned on the target risk:
\begin{equation}
\mathbf{U}=g\!\left(\mathbf{I}_{1:T},\mathbf{M},\mathbf{y},r^{*}\right).
\label{eq:risk-conditioned-motion}
\end{equation}
Second, a multi-view diffusion generator $f(\cdot)$ renders future frames conditioned on the generated motion control:
\begin{equation}
\hat{\mathbf{I}}_{T+1:T+H}=f\!\left(\mathbf{I}_{1:T},\mathbf{M},\mathbf{y},\mathbf{U}\right).
\label{eq:multi-view-generation}
\end{equation}

Our objective is to generate diverse scenarios whose induced risk matches $r^{*}$ while preserving cross-view consistency and temporal plausibility.
\section{Risk-Controlled Motion Generation}
\label{sec:risk_motion}

This section presents per-frame risk computation, risk analysis in typical scenarios, and risk-conditioned motion generation.

\subsection{Per-frame Risk Computation}
\label{subsec:risk_comp}

At each time step $t$, we assume the ego vehicle state $(\mathbf{p}_e^t,\mathbf{v}_e^t)$ and a set of surrounding agents
$\mathcal{A}_t=\{1,\dots,N_t\}$ with states $(\mathbf{p}_i^t,\mathbf{v}_i^t)$ are available from logs or annotations.
We compute a per-agent risk contribution $R_i^t$ and optionally aggregate them into a per-frame scalar risk $r_t$.
Following RiskNet~\cite{liu2026risknet}, we adopt an efficient per-frame, direction-aware risk formulation.

Let the relative displacement and its unit direction be
\begin{equation}
\label{eq:rel_vec}
\mathbf{r}_i^t = \mathbf{p}_i^t - \mathbf{p}_e^t,
\qquad
\hat{\mathbf{r}}_i^t = \frac{\mathbf{r}_i^t}{\lVert \mathbf{r}_i^t \rVert_2 + \epsilon}.
\end{equation}
We use two dot products to indicate whether ego and agent $i$ are moving towards each other:
\begin{equation}
\label{eq:dot_cues}
d_{e\rightarrow i}^t = (\mathbf{v}_e^t)^\top \mathbf{r}_i^t,
\qquad
d_{i\rightarrow e}^t = (\mathbf{v}_i^t)^\top (-\mathbf{r}_i^t).
\end{equation}

Moreover, we map the approaching cues to a discrete interaction weight $\omega_i^t$ that reflects asymmetric driving risk:
\begin{equation}
\label{eq:interaction}
\omega_i^t =
\begin{cases}
\omega_{\mathrm{bi}}, & d_{e\rightarrow i}^t>0~\wedge~ d_{i\rightarrow e}^t>0,\\
\omega_{\mathrm{agent}}, & d_{e\rightarrow i}^t\le 0~\wedge~ d_{i\rightarrow e}^t>0,\\
\omega_{\mathrm{ego}}, & d_{e\rightarrow i}^t>0~\wedge~ d_{i\rightarrow e}^t\le 0,\\
\omega_{\mathrm{away}}, & \text{otherwise}.
\end{cases}
\end{equation}
Typically, $\omega_{\mathrm{bi}}$ is the largest and $\omega_{\mathrm{agent}} > \omega_{\mathrm{ego}}$.

Also, we introduce an agent-type coefficient $\mu_i$ to reflect severity differences across categories
(e.g., heavy trucks vs.\ pedestrians):
\begin{equation}
\label{eq:mass_coeff}
\mu_i = \mathrm{TypeCoeff}(\mathrm{cls}_i),
\end{equation}
where $\mathrm{cls}_i$ is the category label.

To calibrate longitudinal and lateral factors, we define a closing speed along the line of sight
\begin{equation}
\label{eq:closing_speed}
s_i^t = \max\!\Big(0,\; (\mathbf{v}_e^t-\mathbf{v}_i^t)^\top \hat{\mathbf{r}}_i^t \Big),
\end{equation}
and convert it to a longitudinal amplification factor
\begin{equation}
\label{eq:long_factor}
\alpha_i^t = \exp(\kappa\, s_i^t),
\end{equation}
where $\kappa$ controls sensitivity.
To penalize lateral (non-colliding) motion, we compute the sine term using the relative velocity
$\mathbf{v}_{\mathrm{rel}}^t=\mathbf{v}_e^t-\mathbf{v}_i^t$:
\begin{equation}
\label{eq:lateral_sine}
\sin^2\theta_i^t =
\frac{\lVert \mathbf{v}_{\mathrm{rel}}^t \times \hat{\mathbf{r}}_i^t \rVert_2^2}
{\lVert \mathbf{v}_{\mathrm{rel}}^t \rVert_2^2 + \epsilon},
\qquad
\beta_i^t = \exp(-\lambda\, \sin^2\theta_i^t),
\end{equation}
where $\lambda$ controls the lateral attenuation.

Therefore, the final per-agent risk contribution is
\begin{equation}
\label{eq:agent_risk}
R_i^t = K\,C \cdot \frac{\omega_i^t\,\mu_i\,\alpha_i^t\,\beta_i^t}{\lVert \mathbf{r}_i^t \rVert_2 + \epsilon},
\end{equation}
where $K$ and $C$ are calibration constants.
We optionally aggregate per-agent risks into a per-frame scalar risk
\begin{equation}
\label{eq:frame_risk}
r_t = \sum_{i\in\mathcal{A}_t} R_i^t
\quad \text{or} \quad
r_t = \max_{i\in\mathcal{A}_t} R_i^t,
\end{equation}
depending on whether cumulative or maximum risk is used.

\subsection{Risk Mining in Typical Scenarios}
\label{subsec:risk_mining}

Driving risk is often not visually obvious. Many safety-critical situations arise from subtle conflicts among the ego vehicle, surrounding traffic participants, and road geometry, where small timing differences, limited sight distance, or tight lateral clearance can quickly increase crash probability \cite{lan2026latenaux, ling2026vlmped}. To reveal such ``hidden'' hazards, we quantify risk per frame and use the resulting signals to analyze when and why risk accumulates in typical driving scenarios.

\begin{figure}[t]
    \centering
    \includegraphics[width=\columnwidth]{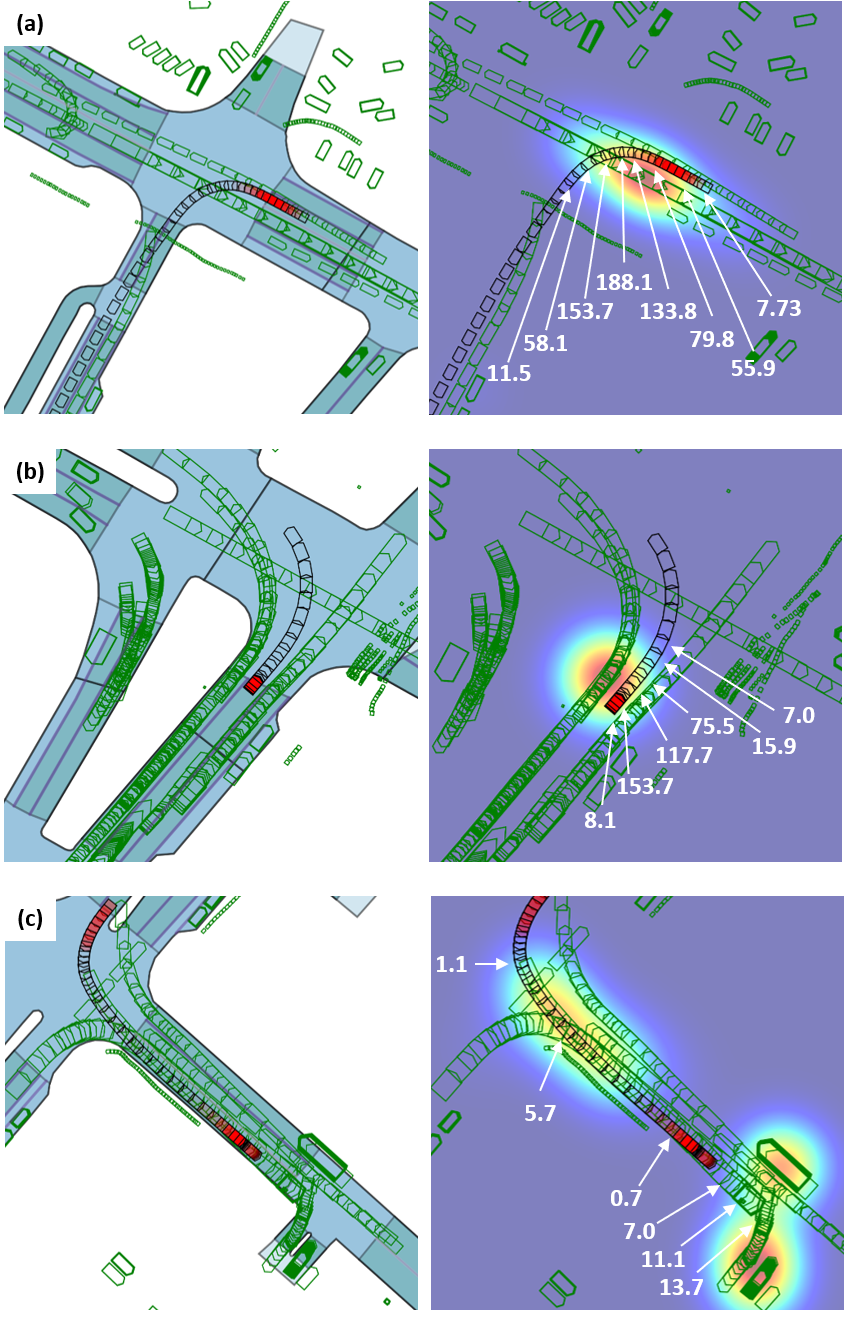}
    \caption{Examples of mined potential risks in typical left-turn scenarios using the proposed per-frame risk quantification.
    The numbers indicate the risk coefficient of the ego vehicle at that location.
    (a) Unprotected left turn at an unsignalized intersection with an oncoming straight-moving vehicle.
    (b) Parallel left turns with tight lateral clearance.
    (c) Left turn followed by a nearby parking-lot exit that creates a local blind spot.}
    \label{fig:risk_examples}
\end{figure}

\begin{figure*}[t]
\centering
\includegraphics[width=\textwidth]{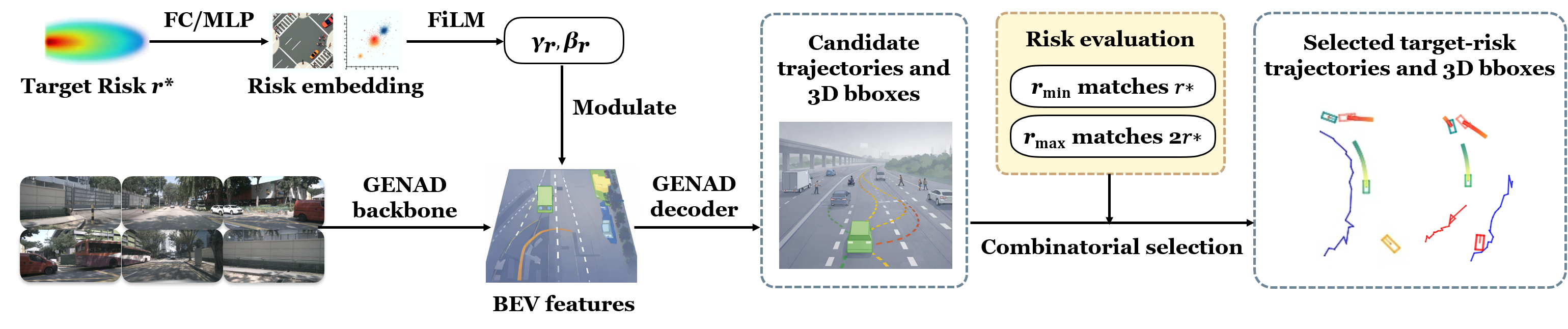}
\caption{Risk-conditioned motion generation. The target risk $r$ is encoded by an MLP and injected into GENAD via FiLM-modulated BEV features. The model predicts trajectories and 3D boxes for the ego vehicle and surrounding agents, followed by risk-aware selection to obtain the final motion set.}
    \label{fig:risk_motion_generation}
\end{figure*}

Large-scale crash statistics further motivate this analysis. NHTSA introduced a pre-crash scenario typology that categorizes hazardous situations into 37 distinct scenarios using the General Estimates System (GES) crash database \cite{najm2007precrash}. A later national analysis reorganized these into nine major groups (rear-end, crossing paths, lane change, road departure, control loss, opposite direction, animal, pedestrian, and pedalcyclist) using 2011--2015 FARS and GES data \cite{swanson2019precrashstats}. Among police-reported crashes where the light vehicle performs the critical action, rear-end and crossing-path conflicts dominate crash exposure (about 34\% and 23\%), followed by lane-change conflicts (about 13\%) and road-departure events (about 11\%). Opposite-direction crashes occur less frequently (about 2\%) but show a much higher fatal-to-all crash ratio (about 3.23\%), indicating that rare scenarios can be disproportionately severe. These findings suggest that risk-aware generation should target not only frequent crash types but also rare, high-severity scenarios often overlooked in manual scenario design.

We mine latent risks in diverse traffic scenarios by computing a per-frame risk signal and analyzing when risk accumulates over time. Figure~\ref{fig:risk_examples} presents three representative left-turn cases. In Fig.~\ref{fig:risk_examples}(a), the ego vehicle performs an unprotected left turn at an unsignalized intersection while an oncoming vehicle proceeds straight, creating a lateral collision threat. The risk rises sharply around the mid-turn phase when the ego vehicle becomes laterally exposed to the oncoming stream. In Fig.~\ref{fig:risk_examples}(b), two vehicles execute parallel left turns with small lateral separation, where slight asynchrony in turning trajectories can induce a sideswipe risk that is hard to anticipate from high-level intent alone. In Fig.~\ref{fig:risk_examples}(c), a parking-lot exit appears immediately after a left turn, forming a localized blind spot, and the risk spikes as the ego vehicle approaches the exit region where cross-traffic may emerge. These examples show that non-intuitive risks are concentrated in specific moments and regions, motivating the use of time-resolved risk as a controllable signal for long-tail motion synthesis and scenario generation.

\subsection{Controllable Motion Generation}
\label{subsec:risk_cond_motion}

We treat risk as a conditioning variable for motion synthesis. Given a target risk $\mathbf{r}^*$ (either a scalar $r^*$ or a profile $\mathbf{r}^*_{1:H}$), our goal is to generate future motion $\mathbf{U}=\{(\mathbf{x}_{1:H}^a,\mathbf{b}_{1:H}^a)\}_{a\in\mathcal{A}}$ such that the induced risk matches the target. We build on a GENAD-style conditional motion generator \cite{zheng2024genad} and introduce an explicit risk-conditioning path.

We then encode the target risk with an MLP $h(\cdot)$ and apply feature-wise linear modulation (FiLM) on the BEV (or latent) feature map
$\mathbf{F}$:
\begin{equation}
\label{eq:film_risk}
\mathbf{e}_r = h(\mathbf{r}^*),\qquad
(\boldsymbol{\gamma}_r,\boldsymbol{\beta}_r) = \psi(\mathbf{e}_r),\qquad
\tilde{\mathbf{F}} = \boldsymbol{\gamma}_r \odot \mathbf{F} + \boldsymbol{\beta}_r,
\end{equation}
where $\psi(\cdot)$ is a linear head and $\odot$ denotes element-wise multiplication.
The modulated feature $\tilde{\mathbf{F}}$ is then used to generate multi-modal future trajectories and 3D boxes.

Let the generator produce $M$ motion hypotheses $\{\hat{\mathbf{U}}^{(m)}\}_{m=1}^M$.
We compute a scalar risk for each mode using the per-frame risk functional in
Eqs.~\eqref{eq:agent_risk}--\eqref{eq:frame_risk}, denoted by $\mathcal{R}(\hat{\mathbf{U}}^{(m)})$.
We enforce a lower-bound matching and an upper-bound capability constraint:
\begin{equation}
\label{eq:risk_minmax}
\begin{aligned}
R_{\min} &= \min_{m} \mathcal{R}\!\left(\hat{\mathbf{U}}^{(m)}\right), \qquad
R_{\max} = \max_{m} \mathcal{R}\!\left(\hat{\mathbf{U}}^{(m)}\right), \\
\mathcal{L}_{\min} &= \big| R_{\min} - r^* \big|, \qquad
\mathcal{L}_{\max} = \big| R_{\max} - \tau r^* \big|.
\end{aligned}
\end{equation}
where $\tau$ is a scaling factor (e.g., $\tau=2$) that encourages the model to retain the capacity to generate higher-risk modes.
The final training loss combines the base GENAD objective $\mathcal{L}_{\mathrm{GENAD}}$ with risk guidance:
\begin{equation}
\label{eq:risk_total}
\mathcal{L}_{\mathrm{risk}}
= \mathcal{L}_{\mathrm{GENAD}}
+ \lambda_{\min}\mathcal{L}_{\min}
+ \lambda_{\max}\mathcal{L}_{\max}.
\end{equation}

Since $r_t$ is computed per frame, we can localize when risk emerges within a scenario by inspecting peaks or sustained segments of $\{r_t\}_{t=1}^H$. This provides interpretable evidence of non-intuitive risk patterns and supports targeted conditioning of motion generation at a desired risk level.

Therefore, as shown in Fig.~\ref{fig:risk_motion_generation}, our module generates risk-conditioned trajectories and 3D bounding boxes for both the ego vehicle and surrounding agents. These outputs serve as the primary motion controls for the subsequent multi-view diffusion model, enabling the generation of driving videos that are consistent with the specified risk level.

\section{Multi-View Diffusion for Driving Scenarios}

\begin{figure*}[t]
\centering
\includegraphics[width=\textwidth]{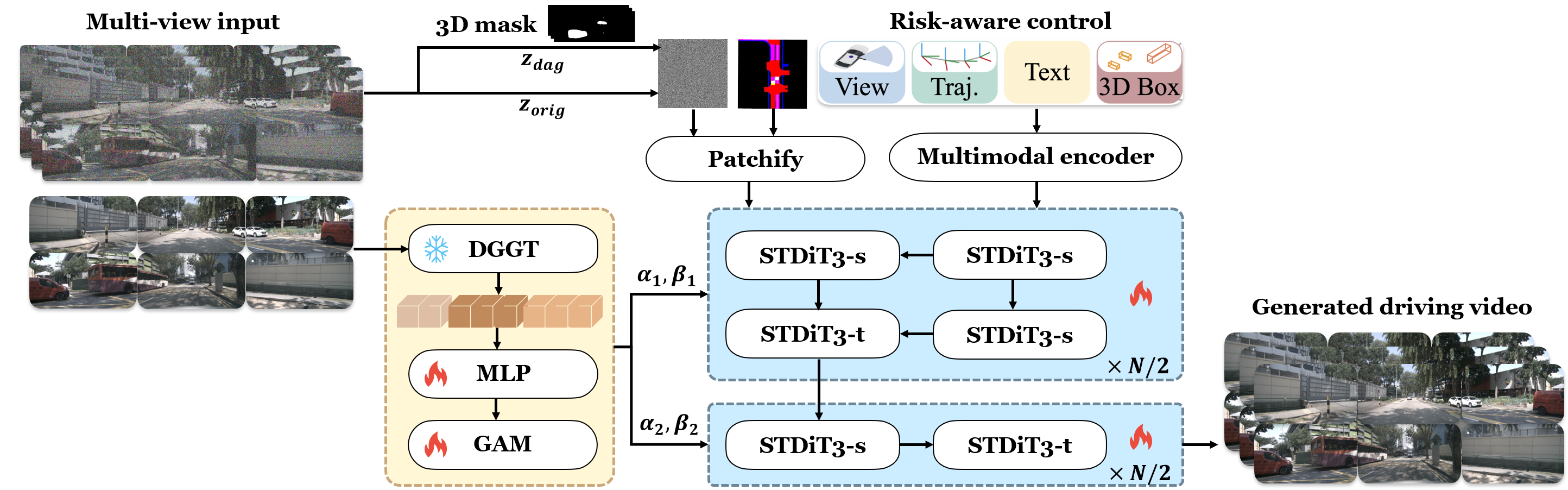}
\caption{
Overview of the proposed \emph{RiskMV-DPO} framework. Given multi-view observations and scenario context, trajectories and 3D bounding boxes generated by the risk control module at a specified risk level are used as structured motion conditions.
A multimodal encoder embeds view tokens and motion cues, which are then injected into a diffusion backbone composed of spatial and temporal STDiT3 blocks.
Region-aware DPO further aligns the generation toward localized dynamic regions, producing temporally coherent multi-view driving videos consistent with the specified risk-conditioned motion.
}
\label{fig:picture2}
\end{figure*}

Figure~\ref{fig:picture2} illustrates the architecture of the proposed framework, termed \emph{RiskMV-DPO}, which integrates diffusion-based generation with multi-view localized preference alignment for risk-aware driving scenario synthesis. Given risk-conditioned motion controls, including trajectories and 3D bounding boxes, the model generates temporally coherent multi-view driving videos. A multimodal encoder fuses view tokens, motion signals, and scenario context, while geometry-aware conditioning and localized DPO guide the generation toward spatially consistent and dynamically plausible driving scenarios.

\subsection{Geometry-Appearance Alignment}
\label{subsec:geo_app_align}

Multi-view diffusion models can produce visually plausible frames, but they often lack a strong geometric prior and may break metric-level cross-view consistency. We inject compact 3D priors by extracting geometry-aware features from DGGT and its VGGT backbone \cite{chen2025dggt,wang2025vggt}, and aligning them with the diffusion model's appearance features during training.

\paragraph{Geometric feature extraction.}
Given the first-frame multi-view images
$ I \in \mathbb{R}^{B \times N_c \times C \times H \times W}$,
where $B$ is the batch size, $N_c$ is the number of cameras, and $C$ is the number of channels,
we extract patch-wise VGGT features
\begin{equation}
F^{\mathrm{VGGT}} = \Phi_{\mathrm{VGGT}}(I),
\qquad
F^{\mathrm{VGGT}} \in \mathbb{R}^{(B N_c) \times P \times D_{\mathrm{VGGT}}},
\label{eq:vggt_feat}
\end{equation}
where $P$ is the number of patches and $D_{\mathrm{VGGT}}$ is the VGGT feature dimension (e.g., $3072$).
A learnable projection maps VGGT features to the diffusion latent dimension $D$ (e.g., $1152$):
\begin{equation}
F = \mathrm{MLP}_{\mathrm{proj}}(F^{\mathrm{VGGT}}),
\qquad
F \in \mathbb{R}^{(B N_c) \times P \times D}.
\label{eq:vggt_proj}
\end{equation}

\paragraph{Learnable geometric token compression.}
To obtain a fixed number of compact geometric tokens, we introduce learnable queries
$Q \in \mathbb{R}^{N_{\mathrm{tok}} \times D}$ (e.g., $N_{\mathrm{tok}}{=}16$) and compress $F$ by cross-attention:
\begin{equation}
G = \mathrm{Attn}(Q, F, F),
\qquad
G \in \mathbb{R}^{(B N_c) \times N_{\mathrm{tok}} \times D}.
\label{eq:geo_token}
\end{equation}
For classifier-free guidance, we apply conditional dropout by replacing $G$ with a learned null token set
$G_{\varnothing}$ using a Bernoulli mask $m_{\mathrm{drop}} \sim \mathrm{Bernoulli}(1-p_{\mathrm{drop}})$:
\begin{equation}
\tilde{G} = m_{\mathrm{drop}} \, G + (1-m_{\mathrm{drop}})\, G_{\varnothing}.
\label{eq:geo_token_dropout}
\end{equation}

\paragraph{Appearance feature extraction.}
During diffusion training, we extract intermediate token features from the last $K$ backbone layers
(e.g., $K{=}8$). Let $R^{(\ell)} \in \mathbb{R}^{(B N_c)\times N_s \times D}$ be the token features from layer $\ell$,
where $N_s$ is the number of spatial tokens.
We spatially pool tokens within each selected layer and average across layers:
\begin{equation}
r^{(\ell)} = \frac{1}{N_s}\sum_{i=1}^{N_s} R^{(\ell)}_{:,i,:},
\qquad
r = \frac{1}{K}\sum_{\ell} r^{(\ell)},
\label{eq:app_feat}
\end{equation}
where $r \in \mathbb{R}^{(B N_c)\times D}$ is the final appearance feature.

\paragraph{Alignment loss.}
We pool geometric tokens into a single vector $g \in \mathbb{R}^{(B N_c)\times D}$,
normalize both $g$ and $r$, and align them with cosine similarity:
\begin{equation}
g = \frac{1}{N_{\mathrm{tok}}}\sum_{k=1}^{N_{\mathrm{tok}}} \tilde{G}_{:,k,:},
\qquad
\hat{g}=\frac{g}{\lVert g\rVert_2},\;\;
\hat{r}=\frac{r}{\lVert r\rVert_2},
\label{eq:geo_pool_norm}
\end{equation}
\begin{equation}
\mathcal{L}_{\mathrm{align}}
= 1 - \frac{1}{B N_c}\sum_{i=1}^{B N_c} \hat{g}_i^{\top}\hat{r}_i.
\label{eq:align_loss}
\end{equation}
This alignment injects compact DGGT/VGGT-derived geometric priors into diffusion training and improves cross-view consistency.

%-------------------------------------------------------------------------

\subsection{Multi-view Consistent Motion-Aware Masking}
\label{subsec:motion_mask}

In driving videos, most pixels belong to static background, while safety-critical cues are concentrated in dynamic regions such as moving agents and their interactions. Uniform corruption wastes model capacity and weakens learning on motion-dominant areas. We therefore construct a motion-aware mask and enforce multi-view consistency so that corruption focuses on dynamic regions while remaining geometrically aligned across cameras.

\paragraph{Motion-aware mask.}
Let $x \in \mathbb{R}^{B\times N_c \times C \times T \times H \times W}$ denote the video latent (or feature) tensor, where $B$ is batch size, $N_c$ is the number of cameras, $C$ is the channel dimension, and $T$ is the number of frames. We estimate per-frame motion magnitude by temporal differencing and channel averaging:
\begin{equation}
\label{eq:motion_map}
\begin{aligned}
D_{b,c,t} &= \big| x_{b,c,:,t+1} - x_{b,c,:,t} \big|, \\
M^{\mathrm{mot}}_{b,c,t} &= \frac{1}{C}\sum_{j=1}^{C} D_{b,c,t,j}.
\end{aligned}
\end{equation}
We normalize $M^{\mathrm{mot}}$ to $[0,1]$ and optionally apply a soft threshold to suppress weak motion.

\paragraph{Geometric consistency across views.}
Motion cues alone do not guarantee cross-view consistency. We additionally render a geometry-consistent mask by sampling 3D points along agent trajectories (or B\'ezier curves) and projecting them into each camera view using known intrinsics and extrinsics. Let $\mathbf{p}=(x,y)$ be a pixel location and $\mathbf{u}_{bck}=(u^x_{bck},u^y_{bck})$ be the projected center of the $k$-th point for batch $b$ and camera $c$, with visibility indicator $V_{bck}\in\{0,1\}$. We rasterize a Gaussian blob:
\begin{equation}
\label{eq:geo_mask_compact}
M^{\mathrm{geo}}_{bck}(\mathbf{p})
= V_{bck}\,
\exp\!\left(
-\frac{(x-u^{x}_{bck})^{2}}{2\sigma_w^{2}}
-\frac{(y-u^{y}_{bck})^{2}}{2\sigma_h^{2}}
\right).
\end{equation}
We aggregate over sampled points to obtain a per-frame, per-view geometric mask $M^{\mathrm{geo}}_{b,c,t}(\mathbf{p})$.

\paragraph{Mask fusion.}
We fuse motion and geometric cues to form the final corruption mask:
\begin{equation}
\label{eq:mask_fusion}
M = \mathrm{clip}\!\big(M^{\mathrm{geo}} \odot M^{\mathrm{mot}},\,0,\,1\big),
\end{equation}
so that corruption focuses on motion-dominant regions while remaining consistent across cameras.

%-------------------------------------------------------------------------

\subsection{Region-Aware Direct Preference Optimization}
\label{subsec:ra_dpo}

Global preference signals can be ambiguous for driving videos and may underweight dynamic regions that dominate risk. We align the model using localized preference pairs constructed on the masked regions, while keeping the unmasked background unchanged.

\paragraph{Progressive corruption fusion.}
Given a noise level $t$ and clean latent $z_0$, we corrupt only the masked region and keep the unmasked region clean:
\begin{equation}
\label{eq:masked_latent}
z_t^{\mathrm{mask}} = M \odot z_t + (1-M)\odot z_0,
\end{equation}
where $z_t$ is the noised latent at level $t$. To construct localized preference pairs, we use two corruption strengths $t_w < t_l$ on the masked region:
\begin{equation}
\label{eq:progressive_pair}
\begin{aligned}
z_{t_w}^{\mathrm{mask}} &= M \odot z_{t_w} + (1-M)\odot z_0, \\
z_{t_l}^{\mathrm{mask}} &= M \odot z_{t_l} + (1-M)\odot z_0.
\end{aligned}
\end{equation}
$t_w$ yields an easier (preferred) target on the masked region, while $t_l$ yields a harder (dispreferred) one, providing localized supervision without altering global context.

\paragraph{Masked flow-matching proxy.}
We maintain an EMA reference model $v_{\mathrm{ref}}$. Using the negative masked flow-matching residual as a proxy for log-probability, we define
\begin{equation}
\label{eq:masked_fm_proxy}
\begin{aligned}
\log p_\theta(y \mid x) &\approx -\, \mathrm{FM}_\theta(y), \\
\delta_\theta &=
v_\theta(z_t^{\mathrm{mask}}, t, \mathbf{c}) - (\epsilon - z_0), \\
\mathrm{FM}_\theta(y)
&= \frac{1}{\lVert M\rVert_{1}}\,
\big\lVert M \odot \delta_\theta \big\rVert_2^{2},
\end{aligned}
\end{equation}
where $\mathbf{c}$ denotes the conditioning (history, map, optional text, and motion tokens), and $\lVert M\rVert_1$ normalizes by masked area.

\paragraph{Region-aware DPO objective.}
We apply DPO on the localized pair $(y_w,y_l)$:
\begin{equation}
\label{eq:ra_dpo}
\begin{aligned}
\mathcal{L}_{\mathrm{RA\mbox{-}DPO}}
&= -\, w(t)\,
\log \sigma\!\Big(
\beta \big[
\Delta_\theta(y_w) - \Delta_\theta(y_l)
\big]
\Big), \\
\Delta_\theta(y)
&= \log p_\theta(y\mid x) - \log p_{\mathrm{ref}}(y\mid x),
\end{aligned}
\end{equation}
where $\sigma(\cdot)$ is the sigmoid, $\beta$ is the DPO temperature, and $w(t)$ is a noise-adaptive weight. This objective aligns generation with localized preferences in dynamic regions.

\paragraph{Supervised fine-tuning loss.}
We apply supervised training on the masked region using the flow-matching residual.
\begin{equation}
\label{eq:sft_loss}
\begin{aligned}
\mathcal{L}_{\mathrm{SFT}}
&= \frac{1}{\lVert M\rVert_{1}}\,
\big\lVert M \odot \delta_\theta \big\rVert_2^{2}, \\
\delta_\theta
&= v_\theta(z_t^{\mathrm{mask}}, t, \mathbf{c}) - (\epsilon - z_0).
\end{aligned}
\end{equation}

\paragraph{Total objective.}
The complete training objective is
\begin{equation}
\label{eq:total_loss}
\mathcal{L}_{\mathrm{total}}
= \lambda_{\mathrm{SFT}}\,\mathcal{L}_{\mathrm{SFT}}
+ \lambda_{\mathrm{RA}}\,\mathcal{L}_{\mathrm{RA\mbox{-}DPO}}
+ \lambda_{\mathrm{align}}\,\mathcal{L}_{\mathrm{align}}.
\end{equation}
\section{Experiments}
\label{sec:experiments}

\subsection{Experimental Setup}
\label{subsec:exp_setup}

We conduct experiments on the nuScenes dataset~\cite{caesar2020nuscenes} using the MagicDrive-STDiT3 backbone.
We use 700 scenes for training and 150 scenes for evaluation.
All videos are processed at $448\times 840$ resolution per camera, and each clip contains $T{=}16$ frames.
We compare RiskMV-DPO with MagicDriveV2~\cite{gao2025magicdrivev2}, DriveDreamer-2~\cite{zhao2024drivedreamer2}, and Panacea~\cite{wen2024panacea}, and also report ablations of our components.

We report eight metrics that cover visual fidelity, temporal quality, and multi-view geometry consistency.
FID measures overall visual quality, and FVD measures temporal coherence.
We evaluate 3D realism using mAP from a pretrained 3D detector applied to generated videos.
To assess multi-view consistency, we report MV-SSIM computed on overlapping regions between adjacent cameras.
For geometry accuracy, we report Depth AbsRel using VGGT-predicted depth against reference depth.
We further report Foreground FID (Fg-FID) on cropped foreground objects and DINO-FID in DINO feature space for semantic fidelity.

We initialize from the MagicDriveV2 checkpoint~\cite{gao2025magicdrivev2} and train for 20{,}000 steps with AdamW.
We use a learning rate of $2\times 10^{-5}$ and batch size 2 per GPU on 8 GPUs.
We adopt BF16 training and an EMA model with decay $\gamma{=}0.9999$.
The trainable parameters include the VGGT geometry adapter ($\sim$45M), the GAM module ($\sim$12M), and the motion-mask generator ($<1$M), totaling $\sim$58M trainable parameters on top of a $\sim$2.5B diffusion backbone.
All experiments are run on NVIDIA H800-90GB GPUs. 

\begin{table}[t]
\centering
\small
\setlength{\tabcolsep}{5pt}
\begin{tabular}{lccc}
\toprule
Method & FID$\downarrow$ & FVD$\downarrow$ & mAP$\uparrow$ \\
\midrule
DriveDreamer-2 & 25.00 & 105.10 & -- \\
Panacea & 16.96 & 139.00 & -- \\
MagicDriveV2 & 20.91 & 94.84 & 18.17 \\
\textbf{RiskMV-DPO (ours)} & \textbf{15.70} & \textbf{87.65} & \textbf{30.50} \\
\bottomrule
\end{tabular}
\caption{Comparison with representative driving video generators on nuScenes.}
\label{tab:main_results}
\end{table}

\begin{table*}[t]
\centering
\small
\setlength{\tabcolsep}{6pt}
\begin{tabular}{lcccccc}
\toprule
Configuration & FID$\downarrow$ & FVD$\downarrow$ & MV-SSIM$\uparrow$ & Depth AbsRel$\downarrow$ & Fg-FID$\downarrow$ & DINO-FID$\downarrow$ \\
\midrule
(A) Baseline & 20.91 & 94.84 & 0.812 & 0.250 & 35.0 & 28.4 \\
(B) + Random Mask DPO & 20.91 & 93.01 & 0.818 & 0.243 & 33.6 & 27.9 \\
(C) + Motion-Aware Mask & 19.42 & 91.67 & 0.825 & 0.236 & 32.1 & 27.4 \\
(D) + 3D MV Mask & 19.33 & 89.51 & 0.841 & 0.228 & 31.0 & 26.8 \\
(E) + VGGT-GAM (no Align) & 16.79 & 88.70 & 0.848 & 0.205 & 30.2 & 26.3 \\
\textbf{(F) RiskMV-DPO (ours)} & \textbf{15.70} & \textbf{87.65} & \textbf{0.856} & \textbf{0.204} & \textbf{29.5} & \textbf{25.8} \\
\bottomrule
\end{tabular}
\caption{Ablation study on nuScenes.}
\label{tab:ablation}
\end{table*}

\begin{figure*}[t]
    \centering
    \begin{minipage}[t]{0.32\textwidth}
        \centering
        \includegraphics[width=\linewidth]{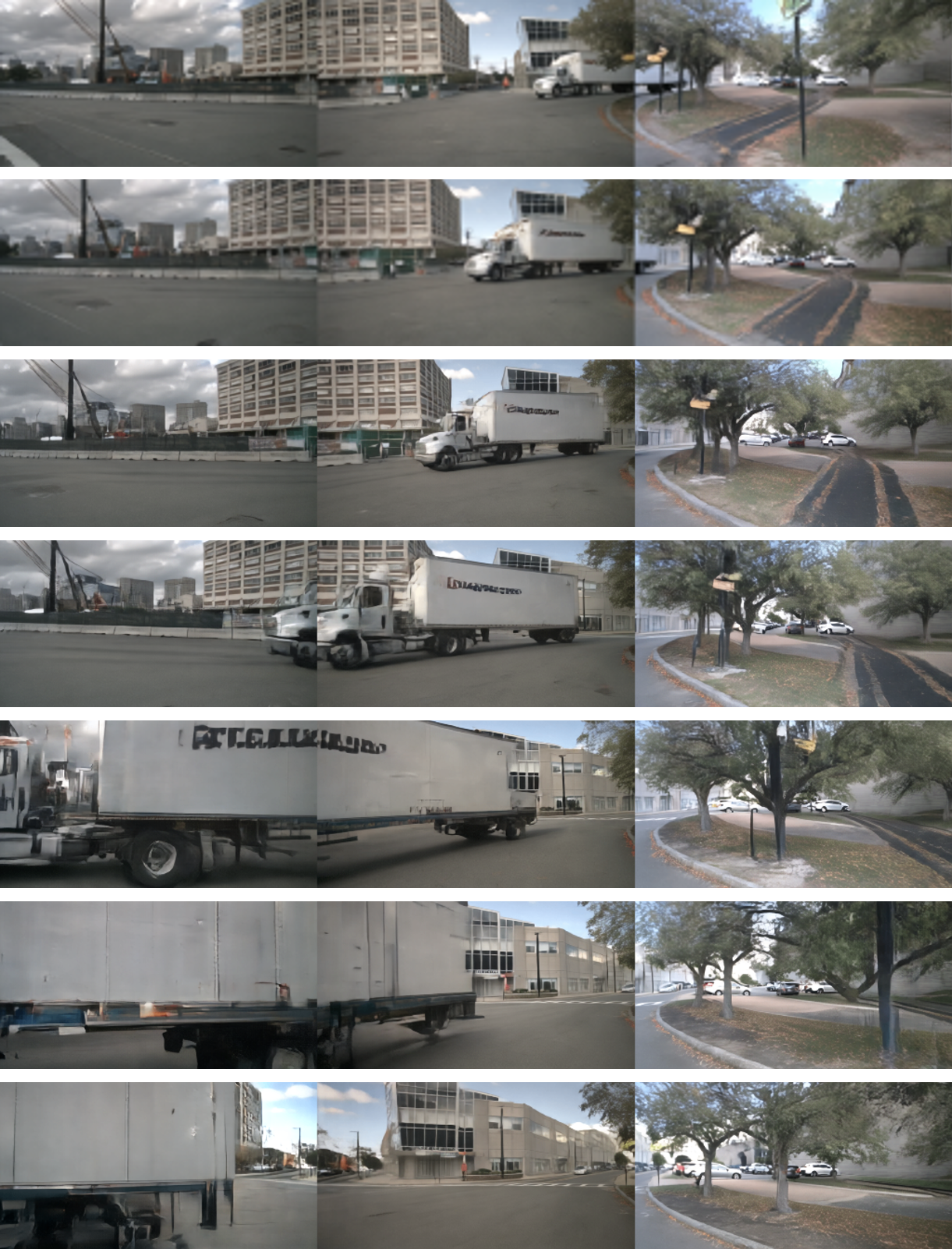}
        \vspace{-1mm}
        \subcaption{0.95 quantile (higher risk)}
        \label{fig:qual_risk_high}
    \end{minipage}\hfill
    \begin{minipage}[t]{0.32\textwidth}
        \centering
        \includegraphics[width=\linewidth]{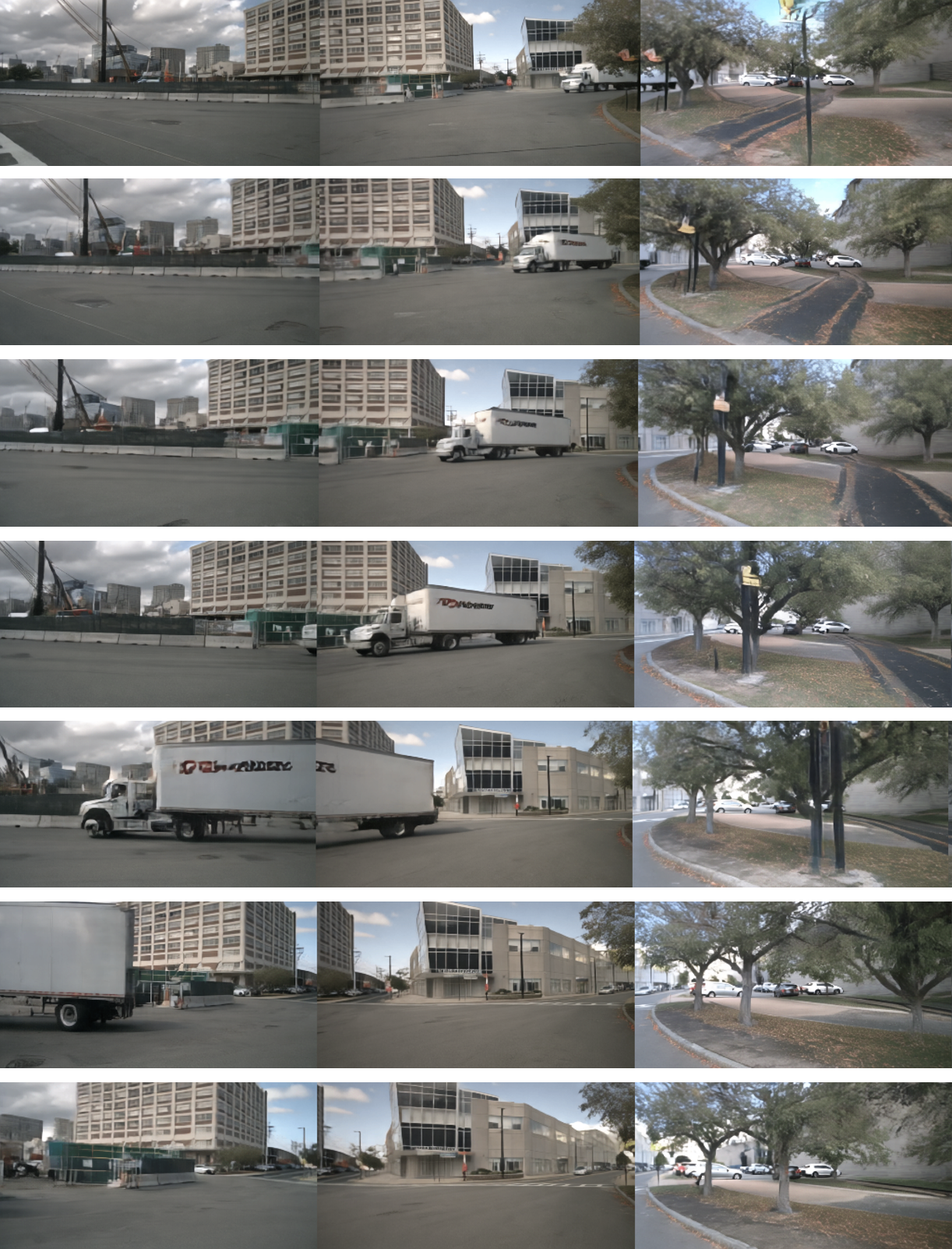}
        \vspace{-1mm}
        \subcaption{0.8 quantile (medium risk)}
        \label{fig:qual_risk_mid}
    \end{minipage}\hfill
    \begin{minipage}[t]{0.32\textwidth}
        \centering
        \includegraphics[width=\linewidth]{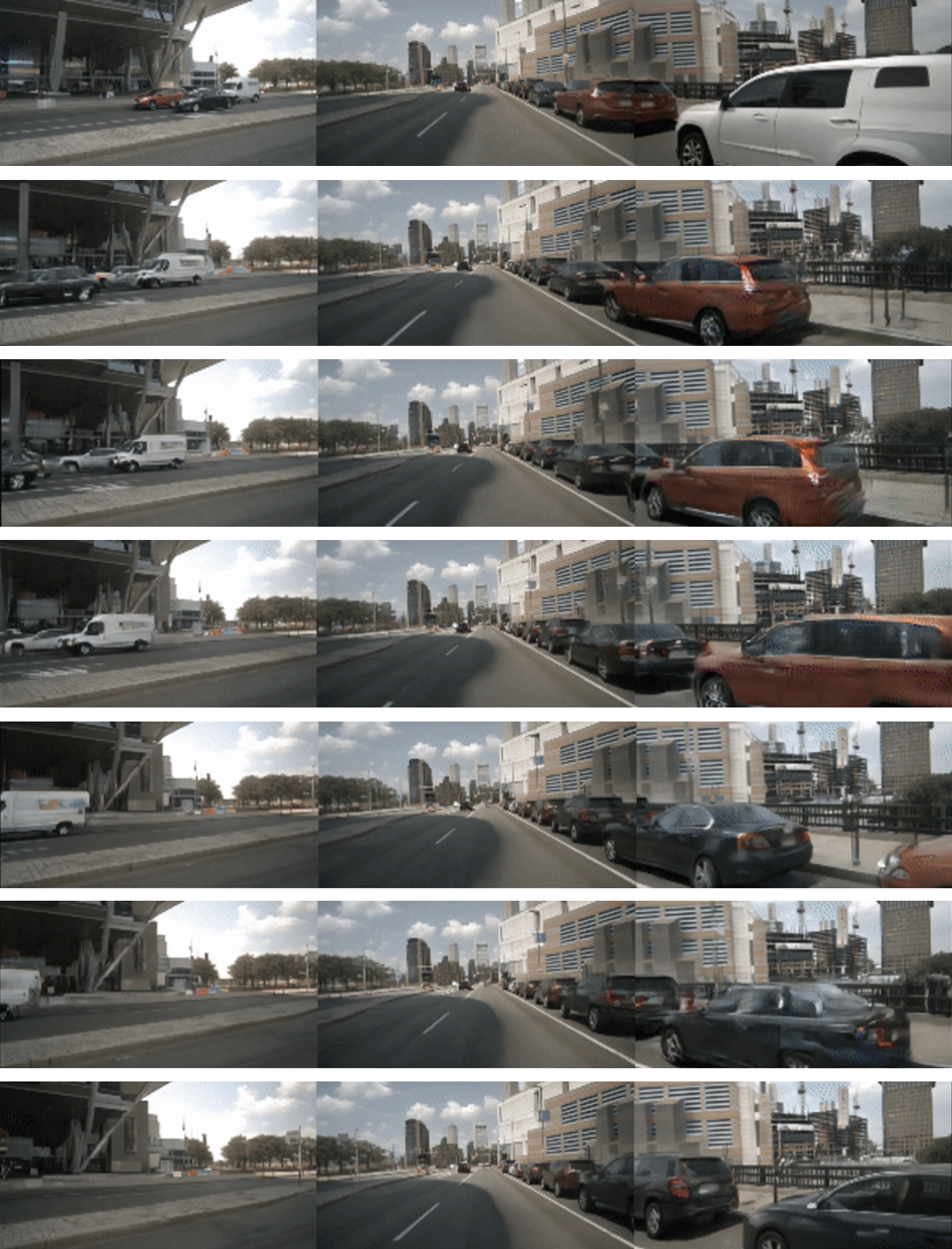}
        \vspace{-1mm}
        \subcaption{0.2 quantile (lower risk)}
        \label{fig:qual_risk_low}
    \end{minipage}
    \caption{Qualitative examples generated by RiskMV-DPO at different target risk quantiles.
    The quantiles indicate the relative position of the induced risk coefficient among all generated scenarios.
    (a) Higher risk: an unsignalized intersection with a fast left-turning heavy truck at very close distance under partial occlusion.
    (b) Medium risk: the same interaction with slower turning and larger clearance, reducing near-collision risk.
    (c) Lower risk: a common straight-driving scenario with no immediate conflict.}
    \label{fig:qual_risk_levels}
\end{figure*}

\subsection{Main Results}
\label{subsec:main_results}

After extensive experiments, we find that the trajectories and 3D bounding boxes generated under a specified risk control lead to synthesized scenarios whose measured risk closely matches the target value. Due to space limitations, we omit the detailed analysis here.

Table~\ref{tab:main_results} reports quantitative results on nuScenes.
RiskMV-DPO achieves consistent improvements over MagicDriveV2.
It reduces FID from 20.91 to 15.70 and improves FVD from 94.84 to 87.65, indicating better visual fidelity and temporal coherence.
It also substantially improves 3D detection mAP from 18.17 to 30.50, suggesting that the generated videos contain more realistic and geometrically consistent 3D cues.
Compared with other recent methods, RiskMV-DPO improves over DriveDreamer-2 (FID 25.00 $\rightarrow$ 15.70) and Panacea (FID 16.96 $\rightarrow$ 15.70).

\subsection{Ablation Study}
\label{subsec:ablation}

Table~\ref{tab:ablation} evaluates the contribution of the key components in our framework.
Starting from the baseline (A), introducing Random Mask DPO (B) slightly improves temporal quality and geometric consistency, but leaves FID unchanged.
Replacing random masks with motion-aware masks (C) yields a clearer gain, improving FID (20.91 $\rightarrow$ 19.42) and MV-SSIM (0.812 $\rightarrow$ 0.825), indicating that focusing training on dynamic regions provides more informative localized preference signals.
Using a 3D multi-view mask (D) further strengthens cross-view consistency, improving MV-SSIM to 0.841 and reducing Depth AbsRel to 0.228.

Next, adding VGGT-GAM without alignment (E) substantially improves geometric accuracy, reducing Depth AbsRel from 0.228 to 0.205, which validates the benefit of injecting geometry priors.
Finally, enabling the geometry-appearance alignment loss (F) achieves the best overall performance, with FID 15.70, MV-SSIM 0.856, and Depth AbsRel 0.204.

\subsection{Qualitative Visualization Across Risk Levels}
\label{subsec:qual_risk_levels}
Figure~\ref{fig:qual_risk_levels} shows qualitative examples generated by RiskMV-DPO at different target risk levels. 
Due to space limitations, we show three representative views (front-left, front, and front-right) and visualize a 7-frame clip by sampling one frame every 5 frames. 
As shown in Fig.~\ref{fig:qual_risk_levels}(a), the higher-quantile scenario depicts the ego vehicle going straight through an unsignalized intersection while encountering a fast left-turning heavy truck at very close distance under partial occlusion, resulting in an imminent collision risk.
In the same scene, the medium-quantile example in Fig.~\ref{fig:qual_risk_levels}(b) shows the truck turning more slowly and keeping a larger clearance to the ego vehicle, which reduces the near-collision risk.
The lower-quantile case in Fig.~\ref{fig:qual_risk_levels}(c) corresponds to a common straight-driving pattern with no immediate conflict, leading to a low estimated risk.

We note that \emph{high}, \emph{medium}, and \emph{low} risk do not have a strict boundary in our setting.
Instead, the reported quantiles indicate the relative position of the induced risk coefficient among all generated scenarios, where 0.95, 0.8, and 0.2 correspond to higher, medium, and lower portions of the risk distribution, respectively, which is apparent on the provided examples. Code and additional qualitative results (including videos) are available at \url{https://github.com/venshow-w/RiskMV-DPO}.
\section{Conclusion and Limitations}

We present RiskMV-DPO, a physically informed pipeline for risk-controllable multi-view driving scenario generation. By treating driving risk as a control signal and coupling risk-conditioned motion synthesis with multi-view diffusion, our framework generates temporally coherent driving scenes at specified risk levels. Experiments on nuScenes demonstrate state-of-the-art performance, improving 3D detection mAP to 30.50 and reducing FID to 15.70. More broadly, our results highlight the value of combining risk-aware modeling with generative world simulation for safety-oriented autonomous driving.

Despite these results, limitations remain. Although the proposed risk modeling captures several typical traffic interactions, many risk patterns in complex real-world environments remain unexplored. In addition, the overall performance depends on the quality of trajectories and 3D bounding boxes generated by the risk control module. Improving the robustness and generality of risk-conditioned motion generation remains an important direction for future work.

\section*{Acknowledgements}
\small This work was supported by the ``U35 Pei Miao Funding Program'' of Changping District, Beijing (Grant No.~CHPU35202504003), National Natural Science Foundation of China (No.~52572334), and the Independent Research Project of the State Key Laboratory of Intelligent Green Vehicle and Mobility, Tsinghua University (No.~ZZ-GG-20250403).

{
    \small
    \bibliographystyle{ieeenat_fullname}
    \bibliography{main}

@String(CVPR= {IEEE Conf. Comput. Vis. Pattern Recog.})

@String(ICLR = {Int. Conf. Learn. Represent.})

@String(AAAI = {AAAI})

@String(CVPR  = {CVPR})

@String(ICLR  = {ICLR})

@inproceedings{kim2021drivegan,
  title={Drivegan: Towards a controllable high-quality neural simulation},
  author={Kim, Seung Wook and Philion, Jonah and Torralba, Antonio and Fidler, Sanja},
  booktitle={Proceedings of the IEEE/CVF Conference on Computer Vision and Pattern Recognition},
  pages={5820--5829},
  year={2021}
}

@article{hu2023gaia,
  title={Gaia-1: A generative world model for autonomous driving},
  author={Hu, Anthony and Russell, Lloyd and Yeo, Hudson and Murez, Zak and Fedoseev, George and Kendall, Alex and Shotton, Jamie and Corrado, Gianluca},
  journal={arXiv preprint arXiv:2309.17080},
  year={2023}
}

@inproceedings{wang2024drivedreamer,
  title={Drivedreamer: Towards real-world-drive world models for autonomous driving},
  author={Wang, Xiaofeng and Zhu, Zheng and Huang, Guan and Chen, Xinze and Zhu, Jiagang and Lu, Jiwen},
  booktitle={European conference on computer vision},
  pages={55--72},
  year={2024},
  organization={Springer}
}

@inproceedings{zhao2025drivedreamer,
  title={Drivedreamer-2: Llm-enhanced world models for diverse driving video generation},
  author={Zhao, Guosheng and Wang, Xiaofeng and Zhu, Zheng and Chen, Xinze and Huang, Guan and Bao, Xiaoyi and Wang, Xingang},
  booktitle={Proceedings of the AAAI Conference on Artificial Intelligence},
  volume={39},
  number={10},
  pages={10412--10420},
  year={2025}
}

@article{gao2023magicdrive,
  title={Magicdrive: Street view generation with diverse 3d geometry control},
  author={Gao, Ruiyuan and Chen, Kai and Xie, Enze and Hong, Lanqing and Li, Zhenguo and Yeung, Dit-Yan and Xu, Qiang},
  journal={arXiv preprint arXiv:2310.02601},
  year={2023}
}

@inproceedings{li2024drivingdiffusion,
  title={DrivingDiffusion: Layout-guided multi-view driving scenarios video generation with latent diffusion model},
  author={Li, Xiaofan and Zhang, Yifu and Ye, Xiaoqing},
  booktitle={European Conference on Computer Vision},
  pages={469--485},
  year={2024},
  organization={Springer}
}

@inproceedings{gao2025magicdrive,
  title={MagicDrive-V2: High-resolution long video generation for autonomous driving with adaptive control},
  author={Gao, Ruiyuan and Chen, Kai and Xiao, Bo and Hong, Lanqing and Li, Zhenguo and Xu, Qiang},
  booktitle={Proceedings of the IEEE/CVF International Conference on Computer Vision},
  pages={28135--28144},
  year={2025}
}

@article{lu2024seeing,
  title={Seeing beyond views: Multi-view driving scene video generation with holistic attention},
  author={Lu, Hannan and Wu, Xiaohe and Wang, Shudong and Qin, Xiameng and Zhang, Xinyu and Han, Junyu and Zuo, Wangmeng and Tao, Ji},
  journal={arXiv preprint arXiv:2412.03520},
  year={2024}
}

@article{xu2023deep,
  title={A deep learning approach for vehicle velocity prediction considering the influence factors of multiple lanes},
  author={Xu, Mingxing and Lin, Hongyi and Liu, Yang},
  journal={Electronic Research Archive},
  volume={31},
  number={1},
  pages={401},
  year={2023},
  publisher={American Institute of Mathematical Sciences}
}

@article{westhofen2023criticality,
  title={Criticality Metrics for Automated Driving: A Review and Suitability Analysis of the State of the Art: L. Westhofen et al.},
  author={Westhofen, Lukas and Neurohr, Christian and Koopmann, Tjark and Butz, Martin and Sch{\"u}tt, Barbara and Utesch, Fabian and Neurohr, Birte and Gutenkunst, Christian and B{\"o}de, Eckard},
  journal={Archives of Computational Methods in Engineering},
  volume={30},
  number={1},
  pages={1--35},
  year={2023},
  publisher={Springer}
}

@article{shalev2017formal,
  title={On a formal model of safe and scalable self-driving cars},
  author={Shalev-Shwartz, Shai and Shammah, Shaked and Shashua, Amnon},
  journal={arXiv preprint arXiv:1708.06374},
  year={2017}
}

@article{liu2026risknet,
  title={RiskNet: interaction-aware risk forecasting for autonomous driving in long-tail scenarios},
  author={Liu, Qichao and Huang, Heye and Zhao, Shiyue and Shi, Lei and Ahn, Soyoung and Li, Xiaopeng},
  journal={Transportation Research Part E: Logistics and Transportation Review},
  volume={205},
  pages={104478},
  year={2026},
  publisher={Elsevier}
}

@article{zhou2026safedrive,
  title={Safedrive: Knowledge-and data-driven risk-sensitive decision-making for autonomous vehicles with large language models},
  author={Zhou, Zhiyuan and Huang, Heye and Li, Boqi and Zhao, Shiyue and Mu, Yao and Wang, Jianqiang},
  journal={Accident Analysis \& Prevention},
  volume={224},
  pages={108299},
  year={2026},
  publisher={Elsevier}
}

@inproceedings{fremont2019scenic,
  title={Scenic: a language for scenario specification and scene generation},
  author={Fremont, Daniel J and Dreossi, Tommaso and Ghosh, Shromona and Yue, Xiangyu and Sangiovanni-Vincentelli, Alberto L and Seshia, Sanjit A},
  booktitle={Proceedings of the 40th ACM SIGPLAN conference on programming language design and implementation},
  pages={63--78},
  year={2019}
}

@inproceedings{dreossi2019verifai,
  title={Verifai: A toolkit for the formal design and analysis of artificial intelligence-based systems},
  author={Dreossi, Tommaso and Fremont, Daniel J and Ghosh, Shromona and Kim, Edward and Ravanbakhsh, Hadi and Vazquez-Chanlatte, Marcell and Seshia, Sanjit A},
  booktitle={International Conference on Computer Aided Verification},
  pages={432--442},
  year={2019},
  organization={Springer}
}

@inproceedings{koren2018adaptive,
  title={Adaptive stress testing for autonomous vehicles},
  author={Koren, Mark and Alsaif, Saud and Lee, Ritchie and Kochenderfer, Mykel J},
  booktitle={2018 IEEE Intelligent Vehicles Symposium (IV)},
  pages={1--7},
  year={2018},
  organization={IEEE}
}

@inproceedings{wu2025drivescape,
  title={Drivescape: High-resolution driving video generation by multi-view feature fusion},
  author={Wu, Wei and Guo, Xi and Tang, Weixuan and Huang, Tingxuan and Wang, Chiyu and Ding, Chenjing},
  booktitle={Proceedings of the Computer Vision and Pattern Recognition Conference},
  pages={17187--17196},
  year={2025}
}

@inproceedings{wallace2024diffusion,
  title={Diffusion model alignment using direct preference optimization},
  author={Wallace, Bram and Dang, Meihua and Rafailov, Rafael and Zhou, Linqi and Lou, Aaron and Purushwalkam, Senthil and Ermon, Stefano and Xiong, Caiming and Joty, Shafiq and Naik, Nikhil},
  booktitle={Proceedings of the IEEE/CVF Conference on Computer Vision and Pattern Recognition},
  pages={8228--8238},
  year={2024}
}

@inproceedings{zhu2025dspo,
  title={DSPO: Direct Score Preference Optimization for Diffusion Model Alignment.},
  author={Zhu, Huaisheng and Xiao, Teng and Honavar, Vasant G},
  year={2025},
  organization={International Conference on Learning Representations (ICLR 2025)}
}

@article{wu2025densedpo,
  title={Densedpo: Fine-grained temporal preference optimization for video diffusion models},
  author={Wu, Ziyi and Kag, Anil and Skorokhodov, Ivan and Menapace, Willi and Mirzaei, Ashkan and Gilitschenski, Igor and Tulyakov, Sergey and Siarohin, Aliaksandr},
  journal={arXiv preprint arXiv:2506.03517},
  year={2025}
}

@article{huang2026mind,
  title={Mind the Generative Details: Direct Localized Detail Preference Optimization for Video Diffusion Models},
  author={Huang, Zitong and Zhang, Kaidong and Ding, Yukang and Gao, Chao and Ding, Rui and Chen, Ying and Zuo, Wangmeng},
  journal={arXiv preprint arXiv:2601.04068},
  year={2026}
}

@inproceedings{wang2025vggt,
  title     = {VGGT: Visual Geometry Grounded Transformer},
  author    = {Wang, Jianyuan and Chen, Minghao and Karaev, Nikita and Vedaldi, Andrea and Rupprecht, Christian and Novotny, David},
  booktitle = {Proceedings of the IEEE/CVF Conference on Computer Vision and Pattern Recognition (CVPR)},
  year      = {2025}
}

@article{chen2025dggt,
  title   = {DGGT: Feedforward 4D Reconstruction of Dynamic Driving Scenes using Unposed Images},
  author  = {Chen, Xiaoxue and Xiong, Ziyi and Chen, Yuantao and Li, Gen and Wang, Nan and Luo, Hongcheng and Chen, Long and Sun, Haiyang and Wang, Bing and Chen, Guang and Ye, Hangjun and Li, Hongyang and Zhang, Ya-Qin and Zhao, Hao},
  journal = {arXiv preprint arXiv:2512.03004},
  year    = {2025}
}

@article{zhang2026world,
  title={World model-based long-tail and scenario-specific generation for autonomous driving},
  author={Zhang, Cong and Wei, Bangyang and Liu, Yang and Labi, Samuel},
  journal={Journal of Intelligent and Connected Vehicles},
  year={2026},
  publisher={清华大学出版社}
}

@article{he2025dual,
  title={Dual-Discriminator Generative Adversarial Network With Long-Tail Feature Capture for Extreme Scenarios in Human--Machine Shared Driving},
  author={He, Xu and Li, Ji and Hu, Chuan and Liu, Mingming and Xu, Hongming},
  journal={IEEE Transactions on Intelligent Transportation Systems},
  year={2025},
  publisher={IEEE}
}

@article{fei2024critical,
  title={Critical roles of control engineering in the development of intelligent and connected vehicles},
  author={Fei, Yang and Shi, Peng and Liu, Yang and Wang, Liang},
  journal={Journal of Intelligent and Connected Vehicles},
  volume={7},
  number={2},
  pages={79--85},
  year={2024},
  publisher={TUP}
}

@article{lin2023generative,
  title={How generative adversarial networks promote the development of intelligent transportation systems: A survey},
  author={Lin, Hongyi and Liu, Yang and Li, Shen and Qu, Xiaobo},
  journal={IEEE/CAA journal of automatica sinica},
  volume={10},
  number={9},
  pages={1781--1796},
  year={2023},
  publisher={IEEE}
}

@article{wang2025safety,
  title={Safety-Critical Scenario Test for Intelligent Vehicles via Hybrid Participation of Natural and Adversarial Agents},
  author={Wang, Yong and Zhang, Daifeng and Li, Yanqiang and Shuai, Liguo and Tang, Zhicheng and Hou, Yuxiang},
  journal={Journal of Intelligent and Connected Vehicles},
  volume={8},
  number={3},
  pages={9210066--1},
  year={2025},
  publisher={TUP}
}

@inproceedings{zhang2024chatscene,
  title={Chatscene: Knowledge-enabled safety-critical scenario generation for autonomous vehicles},
  author={Zhang, Jiawei and Xu, Chejian and Li, Bo},
  booktitle={Proceedings of the IEEE/CVF Conference on Computer Vision and Pattern Recognition},
  pages={15459--15469},
  year={2024}
}

@inproceedings{xu2025diffscene,
  title={Diffscene: Diffusion-based safety-critical scenario generation for autonomous vehicles},
  author={Xu, Chejian and Petiushko, Aleksandr and Zhao, Ding and Li, Bo},
  booktitle={Proceedings of the AAAI conference on artificial intelligence},
  volume={39},
  number={8},
  pages={8797--8805},
  year={2025}
}

@article{cai2026text2scenario,
  title={Text2scenario: Text-driven scenario generation for autonomous driving test},
  author={Cai, Xuan and Bai, Xuesong and Cui, Zhiyong and Xie, Danmu and Fu, Daocheng and Yu, Haiyang and Ren, Yilong},
  journal={Automotive Innovation},
  pages={1--26},
  year={2026},
  publisher={Springer}
}

@article{wang2026vggdrive,
  title={VGGDrive: Empowering Vision-Language Models with Cross-View Geometric Grounding for Autonomous Driving},
  author={Wang, Jie and Li, Guang and Huang, Zhijian and Dang, Chenxu and Ye, Hangjun and Han, Yahong and Chen, Long},
  journal={arXiv preprint arXiv:2602.20794},
  year={2026}
}

@article{zhang2025can,
  title={Can combined virtual-real testing speed up autonomous vehicle testing? Findings from AEB field experiments},
  author={Zhang, Meng and Xu, Jiatong and Gao, Ying and Shen, Dandan and Xu, Zhigang},
  journal={Communications in Transportation Research},
  volume={5},
  pages={100216},
  year={2025},
  publisher={Elsevier}
}

@article{tang2025priorfusion,
  title={PriorFusion: Unified integration of priors for robust road perception in autonomous driving},
  author={Tang, Xuewei and Yang, Mengmeng and Wen, Tuopu and Jia, Peijin and Cui, Le and Luo, Mingshan and Sheng, Kehua and Zhang, Bo and Jiang, Kun and Yang, Diange},
  journal={Communications in Transportation Research},
  volume={5},
  pages={100229},
  year={2025},
  publisher={Elsevier}
}

@inproceedings{zheng2024genad,
  title={Genad: Generative end-to-end autonomous driving},
  author={Zheng, Wenzhao and Song, Ruiqi and Guo, Xianda and Zhang, Chenming and Chen, Long},
  booktitle={European Conference on Computer Vision},
  pages={87--104},
  year={2024},
  organization={Springer}
}

@techreport{najm2007precrash,
  title        = {Pre-Crash Scenario Typology for Crash Avoidance Research},
  author       = {Najm, Wassim G. and Smith, David L. and Yanagisawa, Mikio},
  institution  = {National Highway Traffic Safety Administration},
  number       = {DOT HS 810 767},
  year         = {2007},
  note         = {Based on 2004 NASS GES crash database.}
}

@techreport{swanson2019precrashstats,
  title        = {Statistics of Light-Vehicle Pre-Crash Scenarios Based on 2011--2015 National Crash Data},
  author       = {Swanson, Elizabeth D. and Foderaro, Frank and Yanagisawa, Mikio and Najm, Wassim G. and Azeredo, Philip},
  institution  = {National Highway Traffic Safety Administration},
  number       = {DOT HS 812 745},
  year         = {2019},
  note         = {Scenario groups and statistics based on 2011--2015 FARS and NASS GES.}
}

@inproceedings{caesar2020nuscenes,
  title     = {nuScenes: A Multimodal Dataset for Autonomous Driving},
  author    = {Caesar, Holger and Bankiti, Varun and Lang, Alex H. and Vora, Sourabh and Liong, Venice Erin and Xu, Qiang and Krishnan, Anush and Pan, Yu and Baldan, Giancarlo and Beijbom, Oscar},
  booktitle = {Proceedings of the IEEE/CVF Conference on Computer Vision and Pattern Recognition (CVPR)},
  year      = {2020}
}

@article{gao2025magicdrivev2,
  title   = {MagicDrive-V2: High-Resolution Long Video Generation for Autonomous Driving with Adaptive Control},
  author  = {Gao, Ruiyuan and Chen, Kai and others},
  journal = {arXiv preprint arXiv:2411.13807},
  year    = {2024}
}

@inproceedings{wen2024panacea,
  title     = {Panacea: Panoramic and Controllable Video Generation for Autonomous Driving},
  author    = {Wen, Yu and others},
  booktitle = {Proceedings of the IEEE/CVF Conference on Computer Vision and Pattern Recognition (CVPR)},
  year      = {2024}
}

@article{zhao2024drivedreamer2,
  title   = {DriveDreamer-2: LLM-Enhanced World Models for Diverse Driving Video Generation},
  author  = {Zhao, Guosheng and Wang, Xiaofeng and Zhu, Zheng and Chen, Xinze and Huang, Guan and Bao, Xiaoyi and Wang, Xingang},
  journal = {arXiv preprint arXiv:2403.06845},
  year    = {2024}
}

@misc{qu2023envisioning,
  title={Envisioning the future of transportation: Inspiration of ChatGPT and large models},
  author={Qu, Xiaobo and Lin, Hongyi and Liu, Yang},
  journal={Communications in Transportation Research},
  volume={3},
  pages={100103},
  year={2023},
  publisher={Elsevier}
}

@article{lin2025big,
  title={Big data-driven advancements and future directions in vehicle perception technologies: From autonomous driving to modular buses},
  author={Lin, Hongyi and Liu, Yang and Wang, Liang and Qu, Xiaobo},
  journal={IEEE Transactions on Big Data},
  volume={11},
  number={3},
  pages={1568--1587},
  year={2025},
  publisher={IEEE}
}

@article{he2025generative,
  title={Generative models for the evolution of transportation systems},
  author={He, Yixu and Lin, Hongyi and Yang, Lan and Liu, Yang},
  journal={Journal of Traffic and Transportation Engineering (English Edition)},
  year={2025},
  publisher={Elsevier}
}

@article{lin2025high,
  title={A high-precision calibration and evaluation method based on binocular cameras and LiDAR for intelligent vehicles},
  author={Lin, Hongyi and Liu, Yang and Wang, Liang and Qu, Xiaobo},
  journal={IEEE Transactions on Vehicular Technology},
  volume={74},
  number={5},
  pages={7404--7415},
  year={2025},
  publisher={IEEE}
}

@article{zhang2026dynamic,
  title={A dynamic prompting and scenario generation method for autonomous driving perception via large-model optimization},
  author={Zhang, Song and Lin, Hongyi and Wang, Mengjie and Wei, Bangyang and Liu, Yang and Qu, Xiaobo},
  journal={Transportation Research Part C: Emerging Technologies},
  volume={188},
  pages={105672},
  year={2026},
  publisher={Elsevier}
}

@article{ling2026vlmped,
  title={VLMPed-CoT: A Large Vision-Language Model with Chain-of-Thought Mechanism for Pedestrian Crossing Intention Prediction},
  author={Ling, Yancheng and Qin, Zhenlin and Wang, Leizhen and Liu, Zhendong and Liu, Yang and Ma, Zhenliang},
  journal={Communications in Transportation Research},
  year={2026},
  publisher={清华大学出版社}
}

@article{lan2026latenaux,
  title={LatenAux: Towards Latency-Aware Trajectory Prediction for Autonomous Driving via Consolidated Auxiliary Learning},
  author={Lan, Zhengxing and Liu, Lingshan and Yu, Haiyang and Ren, Yilong},
  journal={Communications in Transportation Research},
  year={2026},
  publisher={清华大学出版社}
}
}

% main PDF no supplementary
% \input{sec/X_suppl}

\end{document}